\documentclass[]{iFlyBot-VLM}

\usepackage{minitoc}
\usepackage{url}
\usepackage{amssymb}
\usepackage[utf8]{inputenc}
\usepackage{microtype}
\usepackage{booktabs}
\usepackage{pifont} 
\usepackage{multirow}
\usepackage{makecell}
\usepackage{paralist}
\usepackage{xspace}
\usepackage{color}
\usepackage{xcolor}
\usepackage{colortbl}
\usepackage{adjustbox}
\usepackage{hyperref} 
\usepackage[edges]{forest}
\usepackage{tikz} 
\usepackage{caption}
\usepackage{amsfonts}
\usepackage[toc,page,header]{appendix}
\usepackage[utf8]{inputenc}
\usepackage{bbm}
\usepackage{enumitem}

\definecolor{seedc}{RGB}{7, 92, 173}

\newcommand{\name}[1]{iFlyBot-VLM}

\newcommand{\hardware}[1]{ByteMini}

\renewcommand{\paragraph}[1]{\vspace{0.1em}\noindent\textbf{#1}}

\title{iFlyBot-VLM Technical Report}

\author[1,\dag]{Xin Nie} 
\author[1,\dag]{Zhiyuan Cheng}
\author[1,\dag]{Yuan Zhang}
\author[2]{Chao Ji}
\author[1]{Jiajia Wu}
\author[1]{Yuhan Zhang}
\author[1,*]{Jia Pan}

\affiliation[1]{iFlyTek Reasearch and Development Group}
\affiliation[2]{LindenBot}

\abstract{
We introduce \textbf{iFlyBot-VLM}, a general-purpose Vision-Language Model (VLM) used to improved the domain of Embodied Intelligence. The central objective of \textbf{iFlyBot-VLM} is to bridge the cross-modal semantic gap between high-dimensional environmental perception and low-level robotic motion control. To this end, the model abstracts complex visual and spatial information into a body-agnostic and transferable "Operational Language", thereby enabling seamless perception–action closed-loop coordination across diverse robotic platforms. The architecture of \textbf{iFlyBot-VLM} is systematically designed to realize four key functional capabilities essential for embodied intelligence: 1) \textbf{Spatial Understanding and Metric Reasoning}; 2) \textbf{Interactive Target Grounding}; 3) \textbf{Action Abstraction and Control Parameter Generation}; 4) \textbf{Task Planning and Skill Sequencing}; We envision \textbf{iFlyBot-VLM} as a scalable and generalizable foundation model for embodied AI, facilitating the progression from specialized task-oriented systems toward generalist, cognitively capable agents.We conducted evaluations on 10 current mainstream embodied intelligence-related VLM (Visual-Language Model) benchmark datasets, such as Blink and Where2Place, and achieved optimal performance while preserving the model's general capabilities. We will publicly release both the training data and model weights to foster further research and development in the field of Embodied Intelligence.
}

\date{\today}
\correspondence{\email{chaoji@iflytek.com,yuanzhang10@iflytek.com}}

\checkdata[Project Page]{\url{https://xuwenjie401.github.io/iFlyBot-VLA.github.io/}}

\begin{document}
\maketitle

\section{Introduction}
\label{sect:intro}

In recent years, with the rapid development of LLM(Large Language Model) and large-scale pre-training technologies, the field of multimodal artificial intelligence has achieved milestone progress\cite{yi2025survey, naveed2025comprehensive}. In particular, Vision-Language Models (VLMs), leveraging their exceptional capabilities in image understanding, text reasoning, and cross-modal knowledge fusion, have become the core driving force behind general perceptual intelligence. These models have demonstrated performance surpassing traditional methods in various general perceptual tasks such as image-text recognition, visual question answering, and content generation. They have also partially integrated into human production and daily life, significantly expanding the application boundaries of AI.

Against this backdrop, researchers have naturally introduced the powerful semantic understanding and reasoning capabilities of VLMs into the fields of Embodied AI and robotics. The goal is to build Vision-Language-Action (VLA) models that can understand complex natural language instructions and execute physical tasks\cite{kawaharazuka2025vision, shao2025large}. Some cutting-edge works, such as $\pi_{0}$\cite{black2024pi_0} and $\pi_{0.5}$\cite{intelligence2025pi_}, use VLMs as key modules to facilitate action generation, painting a promising prospect for the realization of general-purpose embodied agents. However, transferring the excellent performance of VLMs from the digital world to the physical world faces fundamental challenges. These VLA models often rely on massive, highly specialized training data to master a specific skill, which limits their application in diverse and open-ended environments. More seriously, once the operating environment, object posture, or background layout changes, their performance drops sharply. This phenomenon, known as "action memorization," makes them vulnerable to visual changes and results in extremely weak generalization ability. 

On the one hand,This phenomenon is related to model architecture design, and on the other hand, is the huge gap between the current capabilities of VLMs and the capabilities required for embodied intelligence. As is well-known, robots, as intelligent agents operating in the physical world, their physical interactions involve not only digital information flows (such as images and text instructions) but also real-time processing of feedback from the physical world. This feedback includes self-state, fine-grained interactions, and complex dynamic and geometric constraints. The richness and uncertainty of the physical world lead to inherent flaws in the feature space constructed by VLMs in pure digital tasks. When dealing with 3D geometry and interaction relationships unique to the physical world, this feature space often cannot be directly and effectively mapped to robust physical action generation. Specifically, it is difficult for the model to correctly judge the front-back and up-down positions of objects from 2D images alone, to clearly distinguish environmental boundaries such as tabletops and walls, and even harder to accurately estimate the relative distance between objects—let alone plan collision-free operation trajectories. This indicates that existing VLMs cannot effectively extract high-dimensional physical features that are independent of the agent itself and essential for interacting with the physical world. Yet these capabilities are precisely the prerequisites for the generalization of robot manipulation.

\begin{figure}[!t]
    \centering
    \vspace{-0.5cm}
    \includegraphics[width=0.86\linewidth]{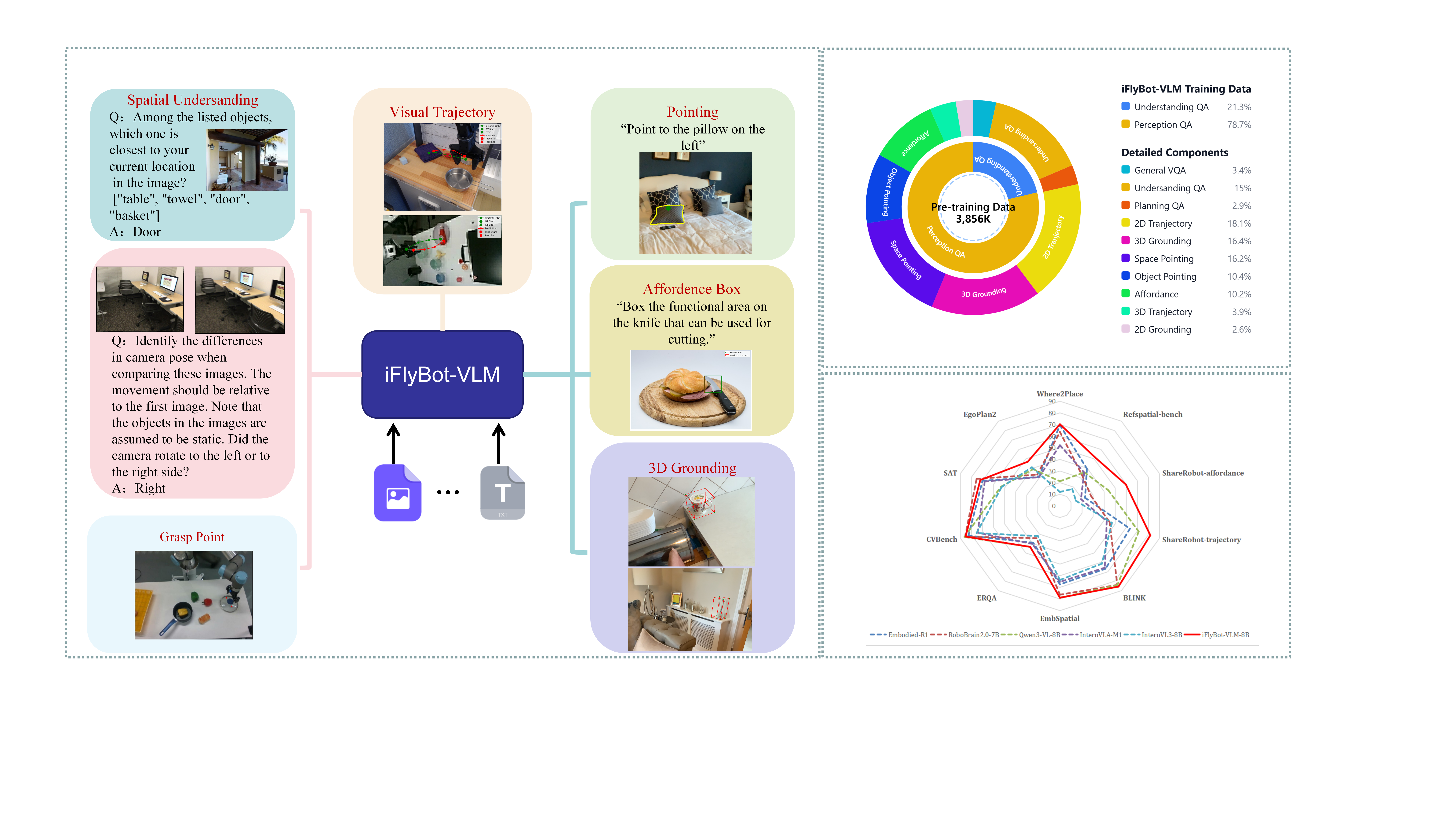}
    \vspace{-0.1cm}
    \caption{\textbf{Overview.}
        The model possesses capabilities in spatial object pointing, 2D trajectories, affordance regions, 3D bounding boxes (3Dbox), object grasping poses, object counting, spatial relationship judgment, multi-image mapping, and task planning. Additionally, it retains excellent multimodal capabilities such as caption generation, Grounding, and Optical Character Recognition (OCR). Moreover, it achieves state-of-the-art (SOTA) performance on multiple evaluation datasets.
    }
    \vspace{-0.3cm}
    \label{fig:teaser}
\end{figure}
Based on this, we propose the iFlyBot-VLM model, which possesses four core capabilities for the embodied domain: spatial understanding and measurement, interactive target localization, action abstraction and control parameter generation, and task planning. These capabilities enable the model to convert environmental information into "operation language" that robots can understand. This "operation language" is transferable and agent-independent, allowing for seamless bridging from perception to action. As illustrated in the Fig,~\ref{fig:teaser}, the four capabilities are as follows:

\textbf{1. Spatial Understanding and Measurement}: Equips the model with the ability to understand the spatial relationships between objects in the environment and estimate their relative positions.

\textbf{2. Interactive Target Localization}: Provides multiple grounding methods, including 2D object detection, 3D object detection, object and spatial pointing, and object affordance region detection.

\textbf{3. Action Abstraction and Control Parameter Generation}: Delivers graspable poses and operation trajectories that are most directly relevant to the manipulation domain.

\textbf{4. Task Planning}: Can predict multi-step future atomic skills based on the current state, laying the foundation for the execution of long-horizon tasks.

This report will elaborate on the model from the perspectives of training data composition, model architecture, and experiments. Chapter 2 details the data composition and data production methods. Chapter 3 presents the model's training framework and training strategies. Chapter 4 describes the evaluation dataset, evaluation metrics, and relevant experiments. Chapter 5 introduces related work, discussing the research progress in embodied foundation models. Chapter 6 provides a summary of this work and an outlook on future research directions.

\section{The Model Architecture}
\label{sec:gr3_model}
\begin{figure}[!t]
    \centering
    \includegraphics[width=\linewidth]{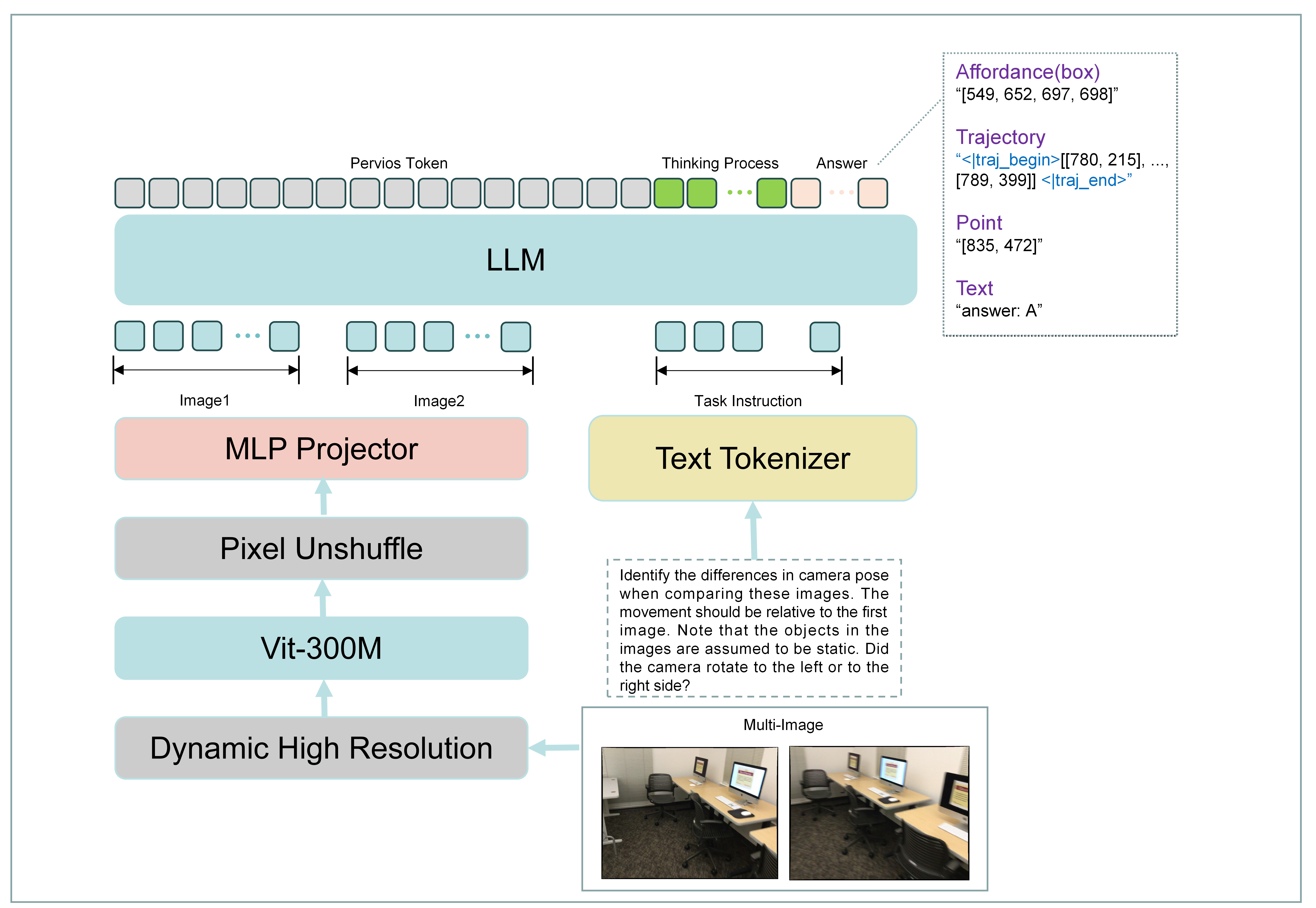}
    \caption{\textbf{The \name{} Model.} \name{} is a three-stage "ViT-Projector-LLM" paradigm from established Vision-Language Models.}
    \label{fig:arch}
\end{figure}

As shown in \figurename{} \ref{fig:arch}, iFlyBot-VLM inherits the robust three-stage "ViT-Projector-LLM" paradigm commonly used in current mainstream VLMs. Specifically, the model achieves efficient feature alignment through a dedicated Vision Transformer (ViT) trained via incremental pre-training, an advanced LLM, and a concise, randomly initialized multi-layer perceptron (MLP Projector). Our model is partially inspired by the architectural design of InternVL3\cite{chen2024internvl}. The LLM and MLP Projector components of our model are initialized using the parameters from InternVL3-8B.

The core innovation of \name{} lies in the improvement of the positional encoding (PE) layer of ViT. Unlike relying solely on the original 448-dimensional positional encoding, this study adopts the bicubic interpolation method to upsample the learned positional embeddings from 448 dimensions to 896 dimensions, thereby forming a new positional encoding mechanism—Dimension-Expanded Position Embedding (DEPE). This innovative method provides a more refined spatial context vector for each visual token, enabling the model to capture more complex positional information and relative spatial relationships without increasing the sequence length. As a result, the model demonstrates superior performance and generalization capabilities in tasks such as fine-grained visual reasoning and precise localization. 
\section{Training Data}
\label{sec:data_training}

\begin{figure}[!t]
    \centering
    \includegraphics[width=0.98\linewidth]{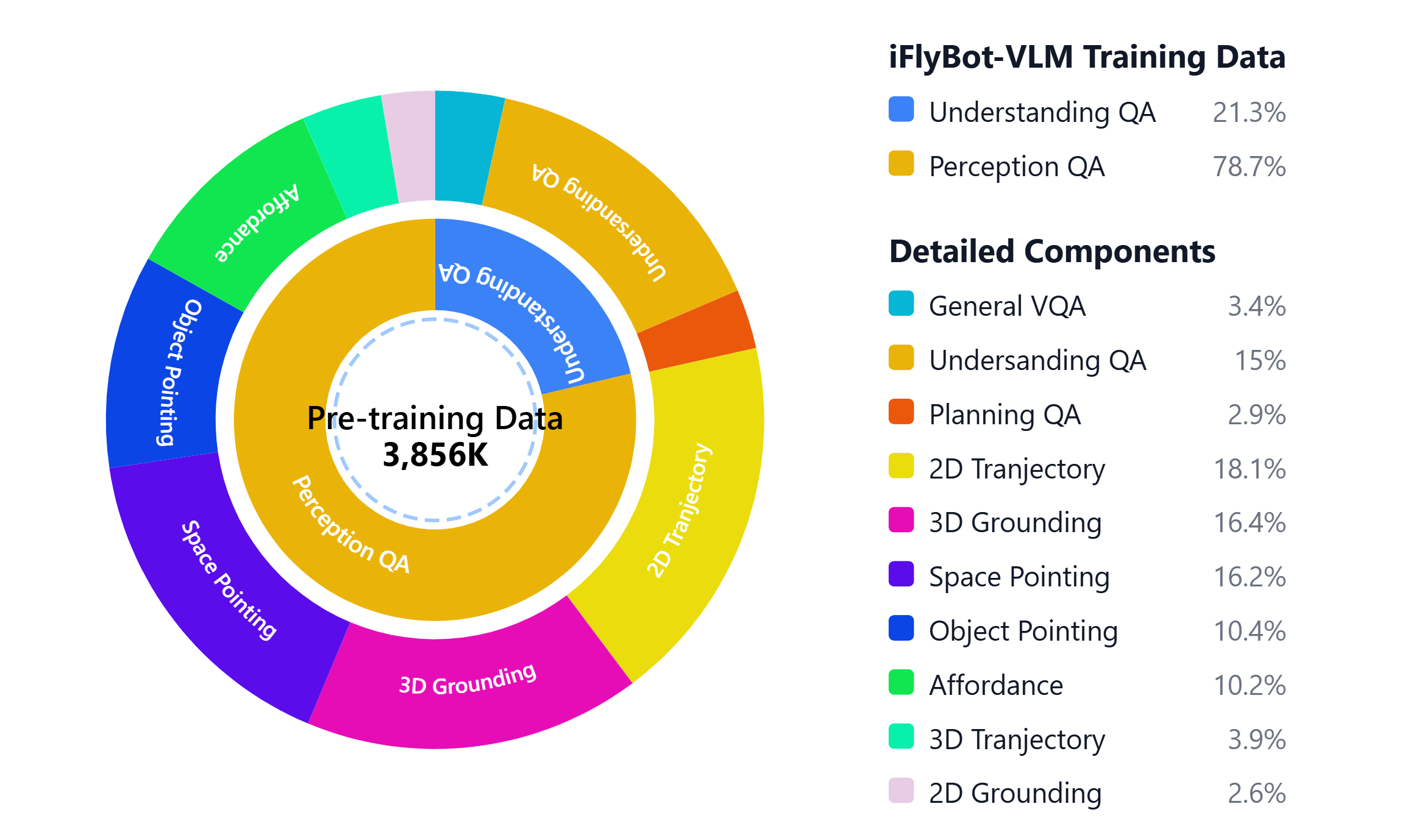}
    \caption{
        Data distribution chart, covering 13 subcategories of data.
    }
    \label{fig:data}
\end{figure}

We train the \name{} model on a mixture of data sources, which is totals approximately 380W samples, covering carefully crafted interactive grounding data, action control parameters, task planning data, and spatial understanding data. Among them, the interactive grounding data amounts to about 215W samples; the action control parameter data—composed of 2D trajectories and object grasping points—totals 73.7W samples; the spatial understanding data sums up to 57.8W samples; and to prevent model forgetting, 13W high-quality general VQA (Visual Question Answering) samples have been added.

\begin{figure}[!t]
    \centering
    \includegraphics[width=0.98\linewidth]{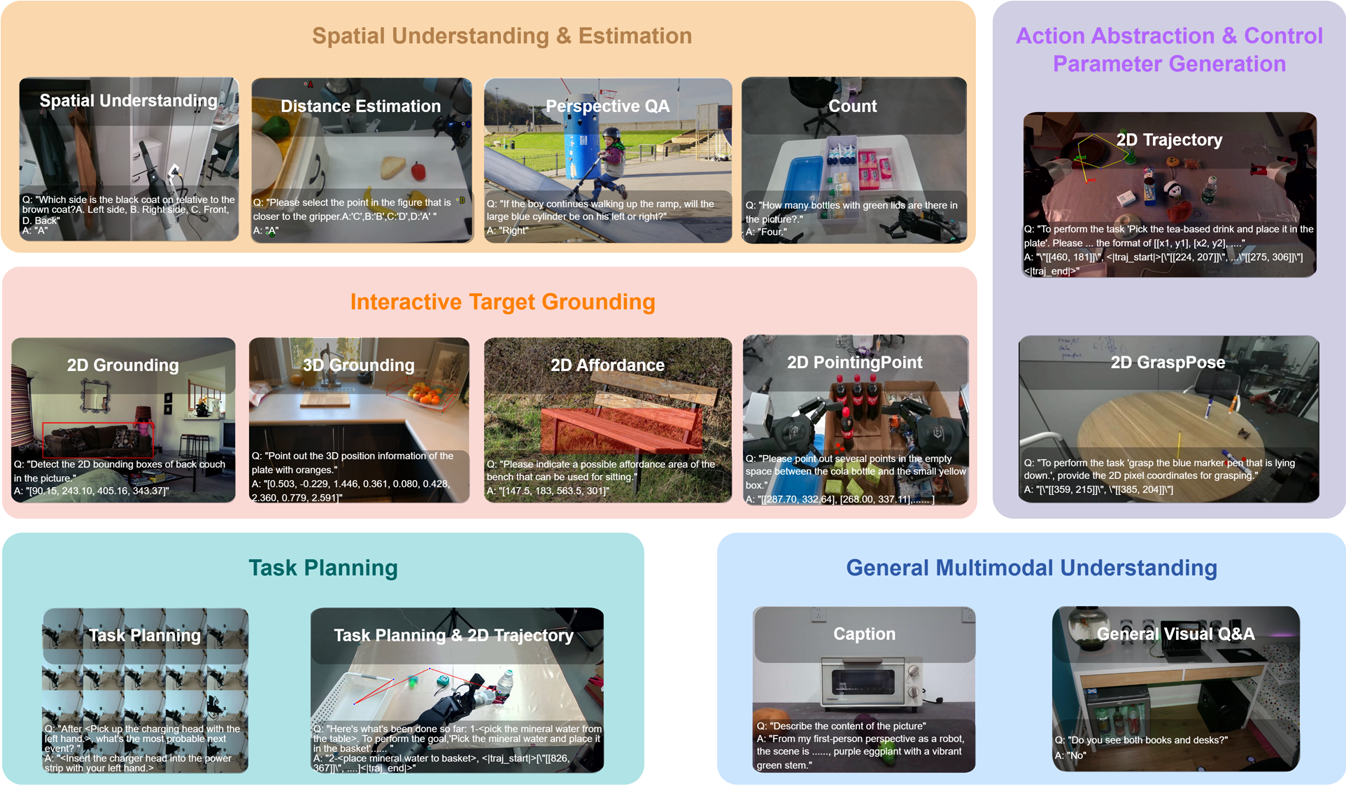}
    \caption{\textbf{Training Data.}The training data presentation covers five categories of data, including General Multimodal Understanding, Action Abstraction \& Control Parameter Generation, Spatial Understanding, Interactive Target Grounding, and Task Planning.}
    \label{fig:data}
\end{figure}

\subsection{\textbf{Spatial Understanding Data}  }
\begin{figure}[!t]
    \centering
    \includegraphics[width=0.98\linewidth]{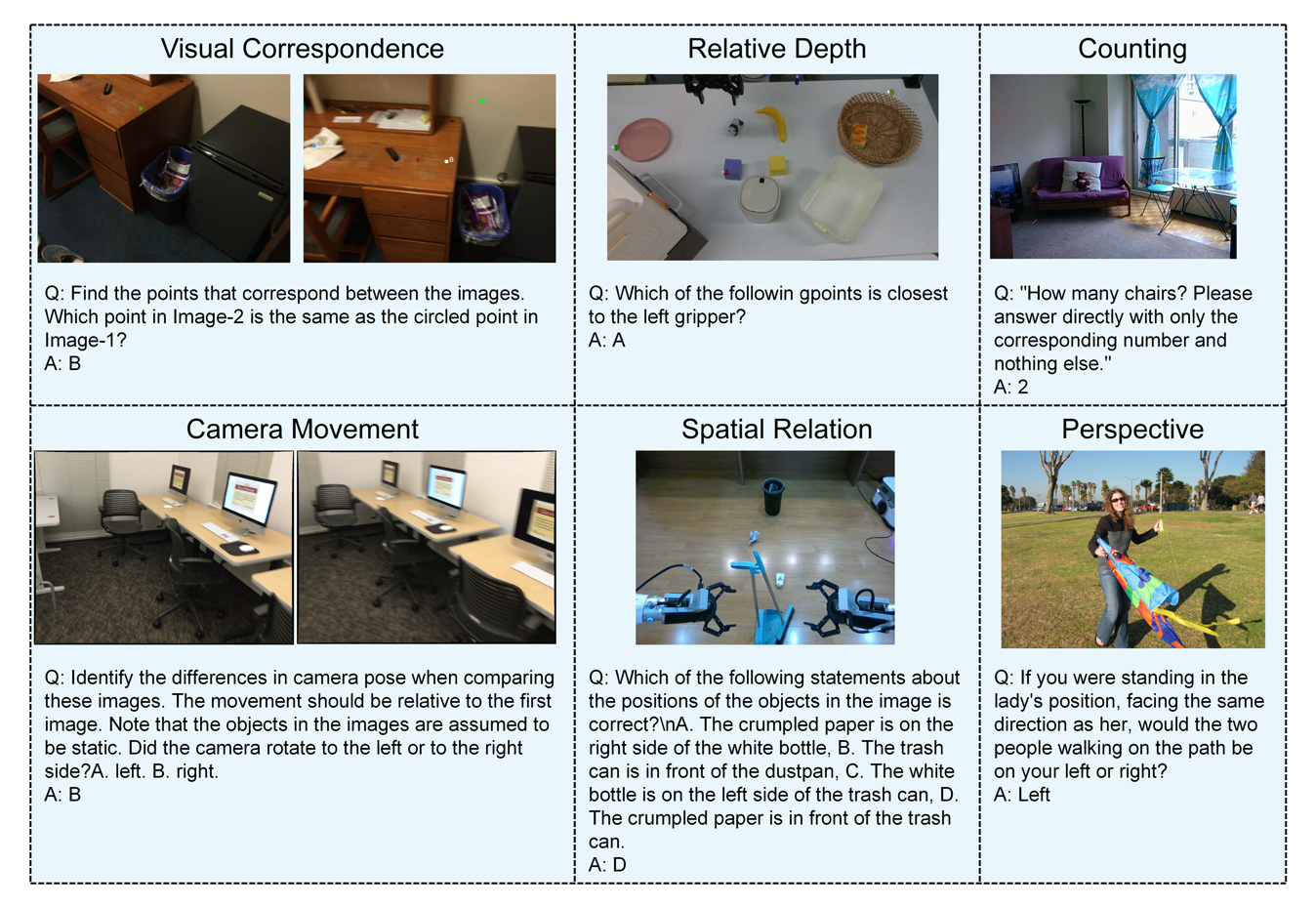}
    \caption{\textbf{Spatial Understanding Data.} This figure presents partial examples of spatial understanding data, including visual correspondence, relative depth, counting, camera movement, spatial relation and perspective}
    \label{fig:saptail-understanding-data}
\end{figure}

Existing Vision-Language Models (VLMs) are capable of performing high-level reasoning and scene interpretation, yet challenges persist in semantic reasoning and scene understanding within robotic manipulation scenarios. As shown in the \figurename{} \ref{fig:saptail-understanding-data} is an overview of our spatial understanding data. The image-text corpora used for VLM pre-training lack high-quality, fine-grained spatial information that serves as a prerequisite for robots to recognize long-tail objects, interpret complex scenes, reason about spatial relationships, and plan physical interactions. Embodied spatial understanding focuses on a robot’s comprehension of object geometry, spatial relationships between objects, object counting, relative depth, camera motion perception, and cross-visual spatial layouts. It covers a range from low-level pattern matching (e.g., visual correspondence) to high-level spatial reasoning (e.g., relative depth).

\subsubsection{Visual Correspondence Data}

This data is designed to enhance the ability of MLLMs to understand and identify the same scene point across different viewpoints, lighting conditions, or timeframes. Using ScanNet\cite{dai2017scannet} and the method outlined in MultiSPA\cite{xu2025multi}, we generated 500K pieces of qualitative data. Given two images and a pixel position in the first image, the model must qualitatively identify the corresponding pixel in the second image that shares the same 3D position. Visual annotations were added to both images: a position point was marked in the first image, and the corresponding position point was marked in the second image via projection. Three additional position points were also annotated to form a multiple-choice question, with the model tasked with selecting the correct annotation.

\subsubsection{Relative Depth Data}

This data aims to improve the geometric understanding capabilities of VLMs. We carefully selected datasets using task data from robotic scenarios (e.g., AgiBotWorld\cite{contributors2024agibotworldrepo}, RoboMind\cite{wu2025robomind}), employed DepthAnything\cite{depthanything} for depth fitting, selected multiple position points in images, and annotated them using depth maps. This process ultimately yielded 34K pieces of high-quality multiple-choice data. To enhance the accuracy of robots in task execution, we designed two types of relative depth data: one for qualitative judgment of relative depth from the current first-person perspective to the camera lens, and another for qualitative judgment of relative depth from the third-person perspective to the robot’s gripper.

\subsubsection{Object Counting Data}

This data is intended to strengthen the detection, recognition, and compositional reasoning capabilities of VLMs for complex scenes where objects may overlap, be occluded, or vary in size and shape. Since pre-trained models already possess a certain level of object counting ability, we used the object counting segment of PixMo\cite{deitke2024molmo} data and TallyQA\cite{acharya2019tallyqa} data, resulting in a total of 84K data points.

\subsubsection{Camera Motion Perception Data}

This data is used to boost the multi-view reasoning capabilities of Multimodal LLMs. Given two images, the model estimates the camera’s relative motion from the first viewpoint to the second, including translation and direction. Using ScanNet\cite{dai2017scannet} and the method described in MultiSPA\cite{xu2025multi}, we annotated rotations as "left" or "right" based on yaw angles, and defined translation directions as "left" or "right" according to the y-axis. To avoid ambiguity during model training, we performed more precise data annotation—for example, removing data where rotation and translation contradicted each other, and appropriately increasing translation distance and rotation angle. Through high-precision annotation, this segment of data totals 61K points.

\subsubsection{Object Spatial Relationship Data}

This data is designed to help VLMs better understand spatial relationships between objects in a scene and improve robots’ ability to interpret complex visual environments. In this dataset, we designed a rich set of spatial relationships, including "left", "right", "in front of", "behind", "above", "under", "between", "nearest", "farthest", "center", and "edge". When designing question-and-answer (Q\&A) data, we created multiple types (e.g., single-choice, multiple-choice, open-ended Q\&A, and image captions) to ensure the model fully learns spatial relationships. Our data sources include the OpenImage\cite{kuznetsova2020open} and VisualGenome\cite{krishna2017visual}. Through high-precision annotation, this dataset contains 406K data points in total.

To further enhance the model’s spatial understanding capabilities, we designed third-person perspective data, which requires judging the relative position of objects from the viewpoint of another observer in the image. An example question is: "If you were the person on the bicycle, facing the same direction as him, would the white truck be in front of or behind you?" This type of data requires prominent targets in the image. Using the YOLO11n\cite{khanam2024yolov11} algorithm, we filtered images meeting the requirements from the VisualGenome\cite{krishna2017visual}, and used Gemini-2.5\cite{comanici2025gemini} to generate the data—resulting in 3K high-precision data points. 

\subsection{\textbf{Spatial Perception Data}  }

In robotic manipulation tasks, in addition to equipping the model with spatial understanding capabilities, precise spatial perception capabilities are also crucial. This involves the position of the manipulated object, the operable regions on the manipulated object, the end-effector posture when the object is manipulated, the positions where the object can be placed, and the trajectory of the end-effector during the object manipulation process.

\subsubsection{2D Grounding Data}
We prepared the iFlyBot-2dGrounding dataset, which includes general scenarios and robotic manipulation scenarios. The general scenario dataset is derived from object detection data in RefCOCO\cite{yu2016modeling} and RoboPoint\cite{yuan2024robopoint}. The objects involved in these datasets include furniture, clothing, food, humans, buildings, plants, animals, etc., and the scenarios cover streets, homes, stadiums, seas, etc. We filtered out data with ambiguous references or meaningless references in the object detection boxes to form the high-quality iFlyBot-2dGrounding-General-250K dataset. However, the scenarios and referenced objects in this data are too generalized, so their role in robotic manipulation tasks is not very direct. Based on this, we supplemented data for robotic manipulation scenarios. This supplementary data is sourced from AgibotWorld\cite{contributors2024agibotworldrepo}, RoboMind\cite{wu2025robomind}, and BridgeData\cite{walke2023bridgedata}. We used Gemini-2.5 and the object lists in the datasets to identify the objects present in the scenarios, then used GroundingDINO\cite{liu2024grounding} to obtain their 2D bounding boxes (2Dbox). After filtering, we constructed the high-quality iFlyBot-2dGrounding-RobotScene-500K dataset.

\subsubsection{2D Pointing Data}

For pointing points, we divided them into two categories: object-pointing datasets and placement-position-pointing datasets. Both categories use 2D points [x, y] as reference results. To expose the model to more diverse data features and distributions, help it learn more general patterns and rules, and thereby improve its generalization ability, we set the reference results as random point sets containing 1 to 9 points. We introduced the RefSpatial\cite{zhou2025roborefer} dataset, which includes 2.5 million high-quality examples and 20 million question-answer (QA) pairs. It contains fine-grained annotations ranging from broad categories to precise spatial references, covering 31 types of spatial relationships. The dataset is composed of multi-source data, integrating 2D web images, 3D embodied videos, and simulated scenarios. This dataset includes both object-pointing data and free-region-pointing data. After filtering the 3D embodied data and simulator data in it, we split the data by type to form two datasets: Iflytek-2dPoints-Location-RefSpatial-270W (for object pointing) and iFlyBot-2dPoints-Placement-RefSpatial-380W (for free-region pointing). In addition, we introduced the RoboPoint\cite{yuan2024robopoint} dataset, which covers 770K pixel point prediction tasks. The answers are mostly represented as lists of 2D points, indicating positions on the image (e.g., task prompts like "locate the object between the markers"). This dataset also includes two categories: object pointing and placement-position pointing. Notably, the images in this dataset typically have a manually labeled reference box for guidance.

For the object-pointing dataset, we specifically introduced RoboReflect-36K\cite{luo2025roboreflect}. Most of the aforementioned datasets focus on indoor home scenes or desktop scenes, lacking robotic arm manipulation scenarios. To address this, we used datasets including AgibotWorld\cite{contributors2024agibotworldrepo}, Droid\cite{khazatsky2024droid}, RoboMind\cite{wu2025robomind}, BridgeData\cite{walke2023bridgedata}, and our self-collected IflytekPickAndPlace data to create the object-pointing dataset for desktop manipulation scenarios: iFlyBot-2dPoints-Object-RobotScene-143K. The creation process was as follows: first, we obtained object bounding boxes using the same method as for iFlyBot-2dGrounding-RobotScene-500K; then, we extracted object contours via SAM (Segment Anything Model); finally, we randomly sampled 1–9 points within these contours. During the creation of this dataset, we defined two types of prompts:
\begin{itemize}
    \item Direct prompts that explicitly specify the object to point to, e.g., "Please indicate the position of the apple";
\end{itemize}
\begin{itemize}
    \item Enhanced prompts that indirectly specify the target object using spatial orientations to further improve the model’s spatial understanding ability, e.g., "Please indicate the third object from the left on the front side of the basket on the table".

\end{itemize}

To further strengthen the model’s understanding of relative distance in spatial comprehension, we introduced the Arkit\cite{baruch2021arkitscenes} dataset. This dataset labels the 3D bounding box coordinates of objects, where the 3D positions and spatial relationships of objects are fixed. We randomly selected objects and used Gemini-2.5 to generate deterministic spatial descriptions between them (e.g., "The coffee table 0.8 meters to the right of the bed"), resulting in the dataset iFlyBot-2dPoints-Location-Arkit-13.9K.

For the object placement position dataset, we also supplemented data for desktop manipulation scenarios. We used robotic manipulation datasets including Droid\cite{khazatsky2024droid}, FSD\cite{yuan2025seeing}, and our self-collected IflytekPickAndPlace data, from which we filtered out only pick-and-place task data. This data typically consists of real-time joint and gripper state variables, and operation videos captured by the head camera. We used the last frame of the fixed head camera sequence as the image for the dataset to be created. For answers composed of point sets, the creation steps were:

    \begin{itemize}
        \item First, we determined the key frames of gripper opening and closing based on changes in gripper state variables.We then used the 2D coordinates of the gripper’s fingertip center at these key frames as a prior;
    \end{itemize}
\begin{itemize}
    \item Next, we obtained the object’s contour via GroundingSAM\cite{ren2024grounded};
\end{itemize}
\begin{itemize}
    \item Finally, we randomly generated point sets within the contour to form the answers.

\end{itemize}

For prompts,we used either the frame before the object was picked up or the first frame of the sequence as the reference image, and used Gemini-2.5 to generate spatial descriptions based on the object at the object point. We defined three types of spatial descriptions:
\begin{itemize}
    \item Between objects;
    \item Between an object and the observer;
    \item Between an object and a specific side of the tabletop.
\end{itemize}
In this way, we constructed the placement-pointing dataset for desktop manipulation scenarios: iFlyBot-2dPoints-Placement-RobotScene-24K. We also reutilized the Arkit\cite{baruch2021arkitscenes} dataset to generate placement-region-pointing data: iFlyBot-2dPoints-Placement-RobotScene-12K.

\subsubsection{2D Affordance Data}

An object’s affordance region refers to the area of the object that can be manipulated. We categorize its references into two types:
\begin{itemize}
    \item Object-centric references: These refer to the regions of the object that can be "manipulated," where "manipulation" includes actions such as sitting on, tapping, picking up, opening, and cutting.
    \item Robot-centric references: These refer to the specific regions of a target object that can be manipulated when the robot intends to operate on that object.
\end{itemize}

We use the coordinate format [x1,y1,x2,y2] to represent these regions.
We first introduced part-level affordance datasets (HANDL\cite{guo2023handal}, InstructPart\cite{wan2024instructpart}, rgbd-part\cite{myers2015affordance}, and iit-aff\cite{Nguyen17}) to create 231.3K data samples. These datasets describe the functional roles of specific parts of objects, with examples such as "Which region of the cabinet can be pulled open?" or "Which part of the axe can be picked up?"

Next, we introduced the PACO\cite{ramanathan2023paco} dataset, which covers 75 object categories, 456 object part categories, and 55 attributes, spanning both image (LVIS\cite{gupta2019lvis}) and video (Ego4D\cite{grauman2022ego4d}) data sources. We filtered this dataset based on criteria such as image resolution and bounding box size, resulting in the iFlyBot-2dAfford-PACO-160K dataset.

Finally, we introduced the affordance dataset AGD20K\cite{luo2022learning}. For images in its training set, we generated prompts using Gemini-2.5, obtained affordance regions via GroundingSAM\cite{ren2024grounded}, and refined these regions through manual adjustments. We then calculated the minimum enclosing rectangles for these regions to form the iFlyBot-2dAfford-AGD20K-12K dataset.

\subsubsection{2D GraspPose Data}

The graspable pose of an object refers to the position and posture of the end-effector when manipulating the object. We use a two-finger gripper as the end-effector, and the grasp pose only needs to be represented by 2 points, i.e., [[x1,y1],[x2,y2]]. The line connecting these two points can be used to indicate the grasp posture.

We used robotic datasets (AgibotWorld\cite{contributors2024agibotworldrepo}, Droid\cite{khazatsky2024droid}, RoboMind\cite{wu2025robomind}, OpenX-Embodiment\cite{o2024open}) to extract key frames of objects before manipulation—specifically, the image frame captured at the mid-moment when the gripper transitions from an open to a closed state. A self-trained object detection model, Gripper-Finger-Detector, was used to detect the two fingertip points, and these two points in the image coordinate system were taken as the grasp points.

Gripper-Finger-Detector was developed through the following process: 20K images were selected and filtered from the aforementioned robotic datasets based on gripper type diversity and scenario diversity, covering 10 gripper styles. After manual annotation, these images were used to train the object detection model. Using this method, we created the dataset iFlyBot-2dGraspPose-OpenSourceRobot-32K from the AgibotWorld\cite{contributors2024agibotworldrepo}, Droid\cite{khazatsky2024droid}, RoboMind\cite{wu2025robomind}, OpenX-Embodiment\cite{o2024open} datasets.

For the self-collected IflytekPickAndPlace data, we have accurate structural dimensions and camera intrinsic/extrinsic parameters. First, we obtained the key grasp frames based on joint state information and gripper information. Next, we calculated the 3D pose of the two gripper fingertips in the camera coordinate system. Finally, we projected these 3D coordinates onto the 2D image via intrinsic parameters to obtain the grasp point coordinates, thus forming the dataset iFlyBot-2dGraspPose-IflytekPickAndPlace-60K. 

\subsubsection{2D Trajectory Data}

We define an object’s operable trajectory as the operation trajectory of the end-effector. We selected 30 sub-actions as atomic skills, including pick, place, open, close, grasp, put, move, pull, push, etc., and defined boundaries for these atomic skills. For example, "pick" refers to the process where the gripper transitions from an empty state (initial state) to grasping an object and then lifting it a certain distance; "place" refers to the process where the gripper (with a load, in a suspended state) places the object and then returns to an empty suspended state.

We introduced the LLARVA\cite{niu2024llarva} dataset and first filtered images with both width and height greater than 224. Additionally, through random sampling checks, we removed images with chaotic scenes or unclear operation subjects. For the point sets in the dataset, we filtered out invalid trajectory data for movement-related instructions (e.g., pick, place): specifically, data where the total path length significantly differed from the total displacement, or where the total path length was excessively short. Since the trajectory points in this dataset were relatively dense, we resampled the trajectories into random sequences of 6 to 8 points by retaining key points. Finally, we modified the dataset’s prompts according to the rules of our defined atomic skills, resulting in the dataset iFlyBot-2dTrajectory-LLARVA-560K.

To expand the scenario coverage, we also created trajectory data using open-source datasets (AgibotWorld\cite{contributors2024agibotworldrepo}, Droid\cite{khazatsky2024droid}, RoboMind\cite{wu2025robomind}, BridgeData\cite{walke2023bridgedata}). The process was as follows: first, we identified key frames of gripper opening/closing based on joint states, and retained 50 sequence frames (including these key frames) for each sequence. Next, we used our self-trained object detector (Gripper-Finger-Detector) to detect the gripper in each sequence frame, and took its central area as trajectory points. Taking Droid data as an example: we first split sequences into sub-task segments based on the key frames of gripper opening/closing; then, using Gemini-2.5\cite{comanici2025gemini}, we split the prompts for these sub-task segments according to our defined list of atomic skills and the sequence descriptions in the original dataset; for each sub-task segment, we extracted the corresponding point set from the original trajectory points; after filtering the point sets based on criteria such as object detection thresholds and the ratio of path length to displacement between the start and end points, we resampled the trajectory sequence for each atomic skill into 6 to 8 random points—with special retention of key frames (the start frame of the sequence and the opening/closing key frames). This process formed the dataset iFlyBot-2dTrajectory-OpenRobotDatasets-181K.

Using the same method as for fingertip points, we obtained real-time pixel coordinates of the gripper center via 3D-to-2D projection from our self-collected IflytekPickAndPlace dataset. We split sequences according to the boundaries of custom atomic skills, extracted trajectory points for each atomic skill, and then resampled them to form the dataset iFlyBot-2dTrajectory-IflytekPickAndPlace-120K. Notably, in addition to trajectories for single atomic skills, we also created two-step trajectory data (e.g., "pick the apple and place it into the basket"). The number of points in such trajectories was limited to 8, including points from key frames (opening/closing frames) and the start frame.

To equip the model with reflective capabilities, we created negative examples for the trajectory data. We applied six types of perturbations to valid trajectories to render them irrational:
\begin{enumerate}
    \item Perturb the first half of the trajectory, so that the start point of the trajectory is not at the gripper;
    \item Perturb the second half of the trajectory, so that the end point of the trajectory is not near the target object;
    \item Perturb the middle part of the trajectory, so that the middle segment passes through obstacles;
    \item Mismatch between text and image: the image lacks the object to be manipulated;
    \item Trajectory missing: the first half of the trajectory is incomplete;
    \item Trajectory missing: the second half of the trajectory is incomplete.
\end{enumerate}

\begin{figure}[!h]
    \centering
    \includegraphics[width=0.98\linewidth]{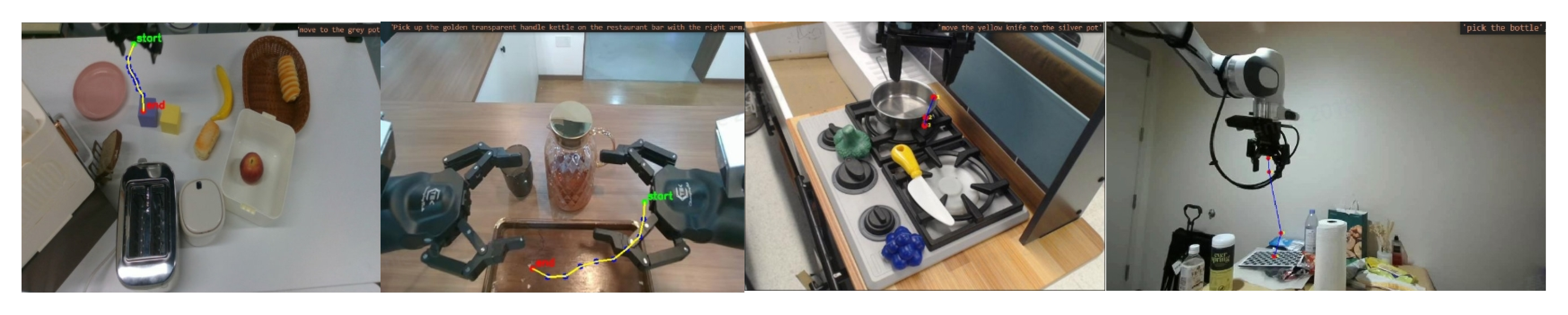}
    \caption{
        Trajectory negative example data.
    }
    \label{fig:data}
\end{figure}

The data structure was designed as follows: the question provides the sequence of trajectory points for the (perturbed) trajectory, and the answer includes a judgment on the trajectory’s rationality and the reason for its irrationality .

In addition, we created a batch of pseudo-reinforcement learning data: the question presents different trajectories, and the answer describes the outcome of following each trajectory (e.g., "the apple was picked up", "the water bottle on the table was touched", "the banana was placed outside the plate"). For this process, we constructed a virtual environment and used human-annotated data as feedback, forming the dataset iFlyBot-2dTrajectory-IflytekPickAndPlace-Reflection-120K.

\subsubsection{3D Grounding Data}

To enhance the model’s spatial measurement capabilities, we created a dataset for 3D bounding boxes (3D boxes), where each 3D box is represented by 9 numerical values: [cx, cy, cz, L, W, H, roll, pitch, yaw]. The first three values (cx, cy, cz) denote the position of the object’s centroid in the camera coordinate system. L, W, and H represent the object’s length, width, and height respectively, with units in meters (m). Roll, pitch, and yaw indicate the object’s orientation, with units in radians (rad).

We introduced the CA-1M\cite{lazarow2025cubify} dataset. Based on the original dataset, we sampled across different shooting scenarios to ensure diversity, used Gemini-2.5\cite{comanici2025gemini} to identify the correct name of each object in the scene, and filtered the dataset based on object size, visibility, and scenarios with mutual occlusion between multiple objects—all to improve the overall quality of the dataset.

For the SUNRGBD\cite{song2015sun}, Objectron\cite{ahmadyan2021objectron}, and ARKitScenes\cite{baruch2021arkitscenes} datasets, we performed filtering based on object visibility and 2D bounding boxes to ensure that all annotated objects in the scene are visually visible.

We noticed that existing open-source datasets contain scenarios with multiple objects that have identical shapes or labels. For such objects, we uniformly added directional ordinal descriptions (e.g., "the first cup from the left") to ensure that each label in the scene refers to a unique object. This enables the model to better locate and reference target objects in complex scenes through directional information.

Furthermore, in addition to direct questions about an object’s 3D box (referred to as Direct-QA), we also created more challenging referential 3D box question-answer (QA) data (referred to as Refer-QA). Instead of directly asking for an object’s 3D box, we ask for the 3D box of an object located in a specific direction relative to another object. This simulates the real-world interaction logic where humans often locate targets through "relative positions" in complex scenes, further enhancing the model’s robustness and practical value when facing multi-object interference and needing to locate objects based on spatial correlations.

\subsection{\textbf{Spatial Perception Data with COT}  }

For complex tasks—such as pointing problems involving involving spatial understanding, part-level affordance judgment, and operation trajectory planning—the core challenge lies in the need for multi-step logical deduction and decomposition of spatial relationships. Directly generating final results can easily lead to jumpy errors or overlooked key constraints. Therefore, we introduce chain-of-thought (CoT) to guide the model in step-by-step reasoning. By breaking down complex tasks into coherent stages such as "identify targets → analyze constraints → derive intermediate intermediate intermediate intermediate steps → verify rationality," the model progresses incrementally based on previous reasoning results. This not only ensures the interpretability of the reasoning process but also reduces omissions caused by one-step generation, ultimately improving the processing accuracy and reliability of complex tasks.Taking trajectory data as an example, we use four steps to guide the model’s reasoning:

\begin{itemize}
    \item Parsing the initial state of objects and the gripper, as well as the placement position;
    \item Clarifying the task and planning obstacle avoidance paths;
    \item Describing the complete trajectory step-by-step (including details of gripper movements);
    \item Drawing the final trajectory conclusion.
\end{itemize}

In this process, the model is prompted to first output the positions of objects and the gripper, which essentially trains the model to master object pointing and target detection capabilities before further deriving trajectories based on established key points. Using this method, we constructed 5K data samples, where the thinking process is enclosed between <think> and </think> as delimiters, followed by the trajectory prediction result.

To enhance the model’s ability to indicate object functional regions, we designed a thinking chain for data augmentation based on the AGD20K\cite{luo2022learning} and InstructPart\cite{wan2024instructpart} datasets. This chain includes identifying user intent, locating object poses, activating the model’s prior knowledge, and positioning functional regions. During the generation of augmented data, we used Grounding DINO to assist in object localization, overcoming potential positioning or reference errors that might occur during generation.

\subsection{\textbf{Task Planning Data}  }

For long-horizon tasks, we often require the model to decompose the task into subtasks for execution, and we incorporate human operation data to improve task accuracy. We utilized the training data from EgoPlan Bench\cite{chen2023egoplan}, where the video data is sourced from Epic-Kitchen\cite{damen2022rescaling} and Ego4D\cite{grauman2022ego4d}. These videos record real first-person perspective data of daily human activities.

Furthermore, to deepen the model’s understanding of task decomposition and how to execute each subtask, we created a hierarchical task trajectory dataset. This dataset integrates both temporal and perceptual data—when outputting subtasks, it simultaneously outputs trajectory results. Based on this, we used our self-collected IflytekPickAndPlace dataset: we selected data with more than 2 steps, split the trajectories using grasp key frames, and used Gemini-2.5\cite{comanici2025gemini} to generate prompts for each segment. Example question descriptions generated include: "For the task of putting all objects on the table into the plate, please predict the remaining steps and the trajectory for each step."

\section{Experiments}
\label{sec:experiments}

To evaluate the model’s performance, we selected commonly used benchmark datasets in the industry.

\subsection{\textbf{Spatial Perceptive Result}  }

\subsubsection{For Pointing Point Evaluation}

First, we chose the Where2Place\cite{yuan2024robopoint} benchmark, an indoor scene dataset focused on object placement regions. It includes 70 seen data samples and 30 unseen data samples. Additionally, we introduced RefSpatial-bench\cite{luo2025roboreflect}, a comprehensive benchmark consisting of 100 object-pointing data samples, 100 placement-region-pointing data samples, and 70 unseen data samples. Both datasets use mask regions to represent ground truth regions.The evaluation score for a single image is calculated as the ratio of the total number of correct predicted points to the total number of predicted points.The evaluation score for a single image is calculated as the ratio of the total number of correct predicted points to the total number of predicted points.Let \( N_{\text{correct}} \) denote the total number of \textbf{correct predicted points} (i.e., the predicted points that fall into the predefined target region, such as a mask area). Let \( N_{\text{total}} \) denote the \textbf{total number of predicted points} generated for the single image.The evaluation score \( S \) for the single image is specifically calculated by the following formula:
\[
S = \frac{N_{\text{correct}}}{N_{\text{total}}}
\]
The total score for the dataset is the average of the sum of scores across all single images.It can be observed that despite being trained using the SFT method, our model outperforms all evaluated models on Where2Place, achieving a score of 70.3. In contrast, RoboBrain2.0-7B scores 63.59, and Embodied-R1 that trained with reinforcement learning scores 69.5. Partial results are shown in \tablename{} \ref{tab:model_performance_selected}.Partial results are presented in Appendix B.On Refspatial-bench,We also ranked first.We achieved a score of 51.5, which is 12 points higher than that of Embodied-R1.

\begin{table}
\captionof{table}{Model Performance on Benchmarks about Spatial Perception}
\label{tab:model_performance_selected}
\begin{center}
\begin{tabular}{lcccc} 
\toprule
\textbf{Models/Metrics} & \textbf{Where2Place} & \textbf{Refspatial-bench} & \textbf{ShareRobot-affordance} & \textbf{ShareRobot-trajectory} \\
\midrule
Embodied-R1\cite{yuan2025embodied} & 69.5 & 39.5 & 22.69 & 63.29 \\
RoboBrain2.0-7B\cite{team2025robobrain} & 63.59 & 32.5 & 28.05 & 44.888 \\
Qwen3-VL-8B\cite{qwen3technicalreport} & 21.2 & 35.75 & 43.72 & 71.21 \\
InternVLA-M1\cite{chen2025internvla} & 52.19 & 35.5 & 18.99 & 42.33 \\
InternVL3-8B\cite{chen2024internvl} & 12 & 18 & 14.45 & 47 \\
\rowcolor{blue!10} 
IflyBot-VLM-8B & 70.23 & 51.5 & 59.61 & 81.74 \\
\bottomrule
\end{tabular}
\end{center}
\end{table}

\subsubsection{For Affordance Dataset Evaluation}

We used the affordance data from Sharerobot-bench\cite{ji2025robobrain} for testing. Before testing, we identified unreasonable aspects in the original ground truth annotations. We manually re-annotated the original images to create Sharerobot-AffordV2-bench. For example, some original annotation boxes were too small, while the affordance region referred to in the text was actually larger (as shown in Figures (a)–(d)): for the text "sit on the couch", the original ground truth region was located in the middle area, which we revised; another example is Figure (d): the text was "Indicate the parts in the figure that can be used for cutting.", but the original annotation did not cover the stem of the knife, so we also corrected this. We also evaluated models such as RoboBrain 2.0 using this revised benchmark.

\begin{figure}[!h]
    \centering
    \includegraphics[width=0.98\linewidth]{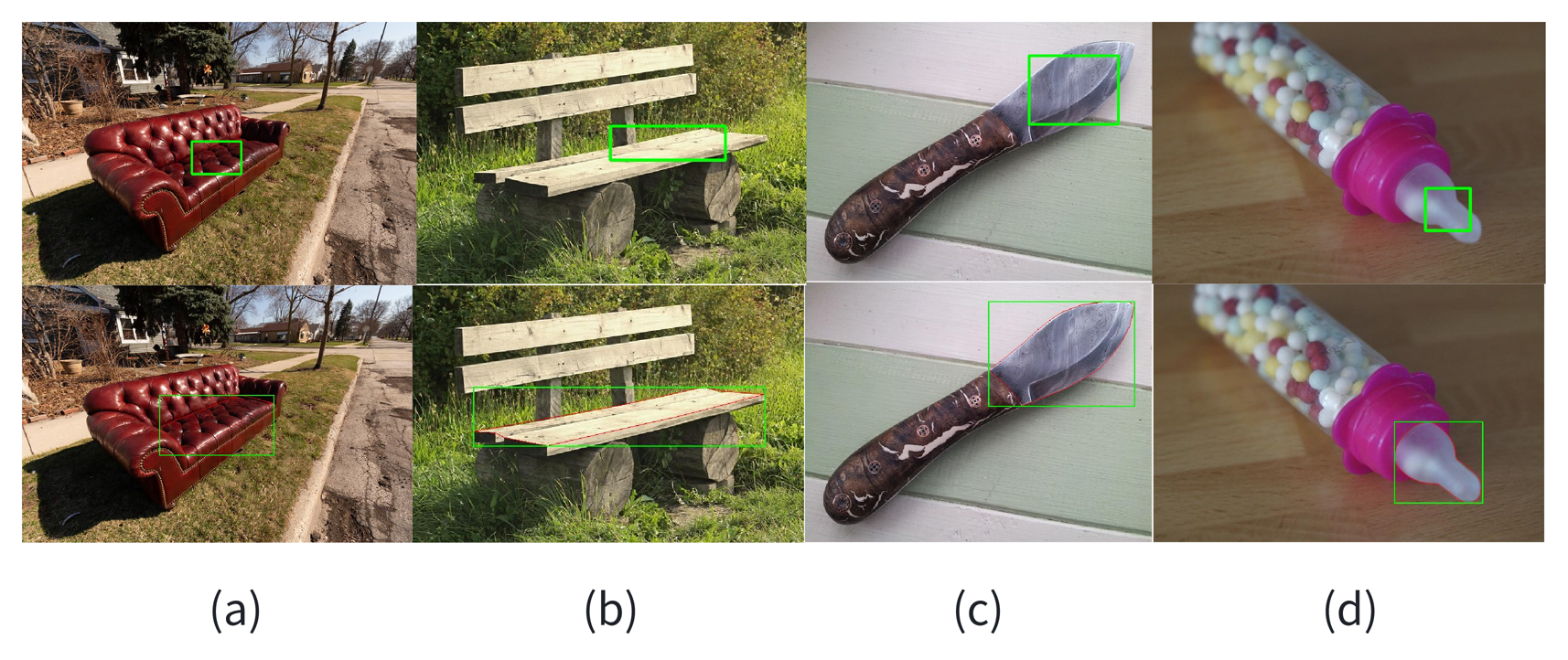}
    \caption{
        Partial results modified based on the ShareRobot-Bench-Affordance benchmark. The upper part shows the original ground truth results, and the lower part shows our annotated results. The area enclosed by the red line is the ground truth mask, and the green-enclosed area is the ground truth bounding box.
    }
    \label{fig:data}
\end{figure}

\subsubsection{For Trajectory Evaluation}

Datasets for trajectory evaluation include the trajectory dataset from Sharerobot-bench and the VABench-V dataset. We selected the trajectory data from Sharerobot-bench for evaluation, which involves judging the similarity between two trajectories. Currently, commonly used metrics include MAE (Mean Absolute Error), RMSE (Root Mean Square Error), and DFD (Dynamic Fréchet Distance). We used the DFD metric to evaluate performance on the trajectory data from Sharerobot-bench.Comparison results show that our model achieved a DFD score of 0.18, while RoboBrain 2.0-7B and RoboBrain 2.0-32B achieved 0.2368 and 0.5512, respectively. The calculation formula is as follows.Notably,We use values ranging from 1 to [the upper limit number] as the final scores, where a higher value indicates better performance.
\begin{equation}
\text{DFD}(P, Q) = \inf_{\gamma, \delta} \max_{t \in [0,1]} \| P(\gamma(t)) - Q(\delta(t)) \|
\end{equation}
where:
\begin{equation*}
\begin{aligned}
P &= \{p_1, p_2, ..., p_K\} \\
Q &= \{q_1, q_2, ..., q_K\}
\end{aligned}
\end{equation*}

To verify the generalization ability of the model, we established a variety of out-of-distribution evaluation scenarios, including: 1) different desktop backgrounds; 2) different containers used as placeable positions; 3) unseen objects; and 4) unseen scenarios, such as office scenarios and factory scenarios.We find that the model could correctly predict trajectories in all cases, and all 8 predicted points are correct.which indicates that the model possesses strong robustness and generalization ability.
\begin{figure}[!h]
    \centering
    \includegraphics[width=0.98\linewidth]{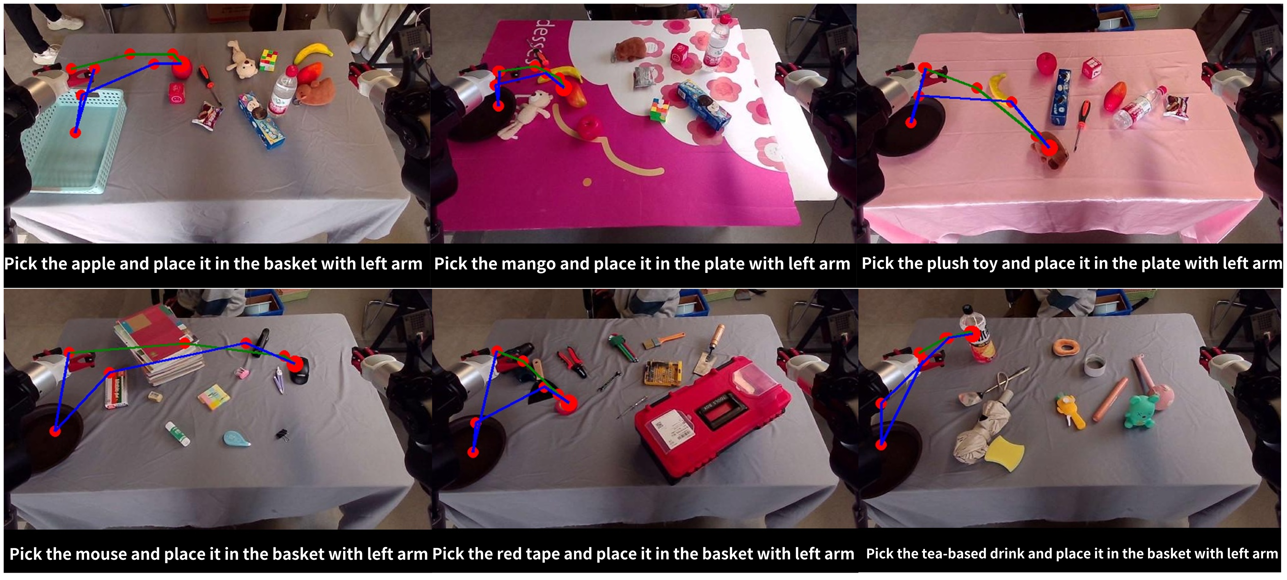}
    \caption{
        The generalization test results under these four unseen scenarios.
    }
    \label{fig:data}
\end{figure}

\subsection{\textbf{Spatial Understanding Result}  }

For spatial understanding and task planning, we conducted comprehensive evaluations of \name{} against existing embodied intelligence models and other MLLMs across six benchmarks. Specifically, the spatial understanding benchmarks include BLINK\cite{fu2024blink}, EmbSpatial\cite{du2024embspatialbenchbenchmarkingspatialunderstanding}, ERQA\cite{team2025gemini}, CVBench\cite{tong2024cambrian}, and SAT\cite{ray2024sat}. The BLINK benchmark comprises 14 sub-tasks, among which we selected the spatially relevant components—Counting, Relative\_Depth, Visual\_Correspondence, and Spatial\_Relation to assess the model’s capability in spatial reasoning.

For task planning, we adopted the EgoPlan2\cite{qiu2024egoplan} benchmark, which provides a convincing measure of embodied task-planning competence. To highlight the superiority of iFlyBot-VLM in spatial understanding, we compared it against state-of-the-art embodied intelligence models that exhibit strong spatial reasoning abilities, including Embodied-R1\cite{yuan2025embodied}, RoboBrain2.0, and InternVLA-M1. In addition, we included comparisons with open-source multimodal large models of comparable scale, such as Qwen3-VL-8B and InternVL3-8B, both of which achieve state-of-the-art performance in general multimodal understanding.

\begin{table}
\captionof{table}{Model Performance on Benchmarks about Spatial}
\label{tab:model_performance_additional}

\begin{center}
\begin{tabular}{lcccccc}
\toprule
\textbf{Models/Metrics} & \textbf{BLINK} & \textbf{EmbSpatial} & \textbf{ERQA} & \textbf{CVBench} & \textbf{SAT} & \textbf{EgoPlan2} \\
\midrule
Embodied-R1\cite{yuan2025embodied} & 66.73 & 67.4 & 39.1 & 82.7 & 70 & 30.82 \\
RoboBrain2.0-7B\cite{team2025robobrain} & 83.95 & 76.32 & 34.25 & 85.75 & 75.33 & 33.23 \\
Qwen3-VL-8B\cite{qwen3technicalreport} & 84.08 & 78.9 & 43.5 & 86.32 & 53.33 & 39.06 \\
InternVLA-M1\cite{chen2025internvla} & 65.65 & 65.27 & 40.35 & 74.41 & 68 & 30.58 \\
InternVL3-8B\cite{chen2024internvl} & 61.36 & 63.63 & 32.08 & 75.36 & 52.67 & 41.03 \\
\rowcolor{blue!10}
IflyBot-VLM-8B & 85.69 & 78.92 & 43.5 & 85.33 & 72 & 47 \\
\bottomrule
\end{tabular}
\end{center}
\end{table}

As shown in \tablename{} \ref{tab:model_performance_additional}, \name{} achieves an accuracy of 85.69\% on the BLINK benchmark, outperforming all other embodied intelligence models and open-source multimodal large models. On the EgoPlan2 benchmark, it attains the best performance with an accuracy of 47\%. On ERQA, iFlyBot-VLM achieves results comparable to Qwen3-VL-8B, while significantly surpassing models such as RoboBrain2.0-7B and Embodied-R1. For CVBench, the model performs on par with Qwen3-VL-8B and RoboBrain2.0-7B, and on SAT, it achieves performance comparable to RoboBrain2.0-7B, while outperforming the current state-of-the-art model Qwen3-VL-8B.

Overall, these results demonstrate that \name{} exhibits strong spatial understanding capabilities, achieving consistently advanced performance across diverse evaluation benchmarks.

\section{Related Work}

\subsection{\textbf{Foundation Models}  }

In recent years, numerous researchers and institutions have focused on the reasoning capabilities of Multimodal Large Language Models (MLLMs), such as Video-llava\cite{lin2023video}, InternVL\cite{chen2024expanding,chen2024internvl}, Gemini 2.5\cite{team2025gemini}, and Qwen3\cite{qwen3technicalreport}. These models have demonstrated outstanding performance in general tasks such as image captioning, video understanding, and multimodal dialog. However, during training, insufficient attention is paid to spatial understanding, perception, and reasoning, resulting in inadequate spatial comprehension abilities. When applied directly to robot manipulation action generation, these models often fail.

Works including SpatialVLM\cite{chen2024spatialvlm}, SpatialRGPT\cite{cheng2024spatialrgpt}, SpatialPIN\cite{ma2024spatialpin} and SpatialBot\cite{cai2025spatialbot} have effectively enhanced environmental perception and interaction capabilities by strengthening models’ spatial understanding. However, they also highlight the immense potential of MLLMs in embodied understanding, achieving promising results on spatial comprehension benchmarks such as BLINK\cite{fu2024blink} and SAT\cite{ray2024sat}.

The aforementioned studies focus on the spatial understanding and reasoning performance of MLLMs. In embodied robot applications, the generalize ability of MLLMs alone is insufficient to achieve effective zero-shot performance on brand-new tasks. Spatial perception capabilities and embodied task planning capabilities are also key factors determining the effectiveness of Vision-Language-Action (VLA) models\cite{yuan2025seeing}.

\subsection{\textbf{Datasets} }

Several works have proposed effective data generation and filtering solutions for spatial reasoning, providing strong support to improve the practical application of MLLMs in robotics and embodied artificial intelligence.Multi-SpatialMLLM\cite{xu2025multi} introduced the MultiSPA dataset, which contains 27 million samples. By integrating depth perception, visual correspondence, and dynamic perception, it enables MLLMs to acquire robust multi-frame spatial understanding capabilities. Meanwhile, it proposed the MultiSPA benchmark, a multi-frame spatial understanding evaluation benchmark.In the interdisciplinary field of spatial reasoning and robotics, InternSpatial\cite{deng2025internspatial} open-sourced a spatial reasoning dataset with 12 million question-answer pairs, covering single-view and multi-view scenarios. Enhances the spatial perception capabilities of robots by supporting supervised fine-tuning (SFT) of VLMs in robotics.

RoboRefer\cite{zhou2025roborefer} proposed the RefSpatial dataset for spatial reference tasks, including 2.5 million samples and 20 million question-answer pairs. It not only provides training support for SFT but also caters to reinforcement fine-tuning (RFT) training.Robo2VLM\cite{chen2025robo2vlm} presented a visual question answering (VQA) dataset generation framework for vision-language models. Enhances VLMs using rich, real multimodal robot trajectory data and constructed the Robo2VLM-1 dataset, which contains 684,710 question-answer pairs.RoboBrain\cite{ji2025robobrain} proposed the ShareRobot dataset, which includes task planning, object affordance, and end-effector trajectories, enhancing the robot’s task execution capabilities.

While the aforementioned datasets have made significant contributions to spatial reasoning in the field of embodied intelligence, they lack comprehensive data specific to the robotics domain. We further refine the robotics domain, enrich the data composition, and further enhance the model’s performance in spatial understanding, perception, and reasoning.

\subsection{\textbf{Embodied Brains} }
Different from general multimodal large models, the goal of embodied VLMs is to equip models with spatial understanding and geometric perception capabilities. Researchers have proposed control parameters directly related to robot manipulation, such as points, trajectories, affordance regions, 2D grounding, and 3D grounding.

\subsubsection{Functional Region Grounding}

Functional region grounding is a crucial perceptual capability for robots to interact with objects. 2D grounding of individual objects cannot enable models to learn how to manipulate objects—especially objects with different part-level functions, such as tools like knives, forks, cups, and axes. Distinguishing the functional attributes of different parts of an object helps robots learn to manipulate objects more effectively. Due to the diversity of interactive objects, this task faces the challenge of establishing clear connections with object parts. \cite{luo2022learning} proposed an affordance grounding task based on an exocentric perspective. Using only affordance labels as supervision, it learns the affordance knowledge of objects and transfers it to egocentric images. It constructed an affordance grounding dataset named AGD20K, which collects and labels over 20,000 images across 36 affordance categories and is a widely used affordance evaluation dataset.

In addition, works such as UAD\cite{tang2025uad}, 2HandedAfforder\cite{heidinger20252handedafforder}, and Afford-X\cite{zhu2025afford} are also dedicated to this field and have proposed various affordance datasets at both the part level and the entire object level—such as Handle, LVIS-Aff, and COCO-Aff—which annotate functional attributes from multiple action dimensions. Affordance-R1\cite{wang2025affordance} built a high-quality reasoning dataset centered on functional affordance, named Reason Aff. Trained via reinforcement learning based on GRPO\cite{guo2025deepseek}, it achieves strong zero-shot generalization capabilities and demonstrates emergent in-test reasoning abilities.

\subsubsection{Points and Trajectories as Bridges}

Many institutions have researched the use of points and trajectory points as bridges between the "brain" and action modules, achieving promising results.RoboPoint\cite{yuan2024robopoint} proposed a vision-language model that predicts the adaptability of key points in images based on language instructions, adapting to embodied tasks.Hamster\cite{li2025hamster} fine-tuned a VLM using trajectory datasets generated from simulator and real robot datasets, and applied these trajectories as a bridge to action generation. It achieved a 20\% success rate improvement in 7 task tests, demonstrating the importance of trajectories as a control parameter. MolmoAct\cite{lee2025molmoact} proposed a dataset containing over 10,000 high-quality robot trajectories from different scenarios and tasks. After training on this dataset, the overall performance improved by an average of 5.5\% compared to the base model.
A0\cite{xu2025a0} captures object-centric spatial affordance by predicting contact points and post-contact trajectories. It was pre-trained on 1 million contact point data and fine-tuned on annotated trajectories. After mapping the inferred trajectory data to 3D space, it achieved state-of-the-art (SOTA) results compared to RDT in wiping tasks.In addition, Magma\cite{lu2022phrase} has also achieved notable results in the use of points.

\subsubsection{Comprehensive Models}

Some comprehensive models, such as Embodied-FSD, leverage the visual understanding capabilities of VLMs. They use data such as spatial affordance boxes/points and visual trajectories to provide expressive and compact spatial information. By introducing the FSD model and the SrCOT method, they achieve spatial understanding and reasoning in embodied tasks, and proposed the FSD dataset. After fine-tuning on some open-source models using this data, the average success rate of task execution increased by more than 10\%. In addition, 2D trajectories inferred based on COT are mapped to 3D space, and complex tasks such as cloth folding are completed using Curobo for operation execution.

RoboBrain enhances robot manipulation capabilities by constructing ShareRobot data and general multimodal data, and adopting a multi-stage training strategy. Embodied-R1\cite{yuan2025embodied} uses "pointing" as an intermediate state to connect high-level vision-language understanding and low-level actions. Through two-stage RFT training, it builds the model’s spatial reasoning capabilities and multi-task embodied pointing capabilities.
RoboBrain-2.0\cite{team2025robobrain} adopts a multi-stage training strategy, enabling the model to support spatial understanding, temporal modeling, and chain-of-thought reasoning in embodied environments. InternVLA-M1\cite{chen2025internvla} was pre-trained on over 2.3 million spatial reasoning data samples, taking spatial localization as a key link in vision-language-action training, highlighting that spatially guided training can scale the embodied capabilities of general robots.

Besides, 3D grounding capability is also an important control parameter for embodied brains, which is more conducive to models learning spatial relationships and sizes between objects. For example, Qwen3-VL and Gemini 2.0 have conducted research in this area. 

These works promote the advancement of embodied foundation models from the 2D world to the 3D world.The aforementioned solutions have made significant breakthroughs in spatial understanding, reasoning, and task planning. However, the continuous improvement of robot embodied tasks requires robots to possess certain general thinking and reasoning capabilities. Based on this, this paper further refines spatial reasoning tasks to enhance the spatial understanding capabilities of MLLMs. Additionally, by constructing rich multimodal data, we improve the spatial reasoning, understanding, and task planning capabilities of MLLMs while equipping them with general image-text reasoning capabilities.
\section{Conlusion \& Future Work}
\label{sec:conclusions}

In this report, we introduce our vision-language foundation model designed for embodied scenarios, and elaborate on the dataset creation method, model architecture, and evaluation approach in detail. We have open-sourced the model checkpoints, datasets, and evaluation tools. Our model possesses spatial understanding and reasoning capabilities, task planning capabilities, as well as spatial perception capabilities including spatial pointing, affordance regions, 2D trajectories, object grasping poses, and 3D bounding boxes (3Dbox). Compared with general multimodal models and the latest industry-leading embodied multimodal models on typical evaluation datasets, our model has achieved outstanding performance.

In future work, we will continue to enhance the model’s spatial understanding and spatial perception capabilities, and densely integrate them into Vision-Language-Action (VLA) models for application. We will focus on the following key directions:
\begin{enumerate}
    \item Incorporating image generation capabilities into the model, and introducing a world model to enable the model to learn future prediction. This will enhance spatial understanding capabilities and yield more accurate 2D and 3D trajectories.
    \item Expanding the model’s self-reflection ability and multimodal input-output capabilities such as scene reconstruction, point clouds, and 3D trajectories.
    \item Applying this model to the VLA framework to improve the model’s generalization ability and step-by-step execution capability for long-horizon tasks.
\end{enumerate}
\label{sec:references}

\bibliographystyle{IEEEtran}  
\bibliography{main}  
\section{Appendix}
\label{sec:appendix}

\subsection{\textbf{Appendix A}  }
\subsubsection{Prompt for Chain-of-Thought (CoT) on Affordance}

\begin{tcolorbox}[
    colframe=blue!30,
    colback=gray!5,
    rounded corners,
    boxrule=1pt,
    breakable, 
]

You are a senior data annotation expert. Your task is to analyze image information based on the given image and User Query, and generate a detailed English thinking process to identify the \textbf{affordance area} of the target object from a robotic perspective that enables the specified task. Please strictly follow the below task requirements and thinking steps for your output.

\section*{Task Requirements}
1. Understand the User Query and image content. The User Query will clearly specify an object and a task that can be completed. The image provides visual information, with a red rectangular box marking the corresponding affordance area.  
2. Develop a logical English thinking process. This process must mimic the "reasoning" of a Vision-Language Model (VLM): starting from the image content and user query, and gradually deducing which area is the affordance area for completing the specified task. Your thinking process must align with the position of the affordance area and the object mentioned in the User Query.  
3. Adhere to key constraints: Do not reveal the fact that the image has manual annotations. Your thinking process must be independent and derived from "observation" and "understanding" of the image itself. You must not state phrases like "there is already a red box on the image" or "this box is pre-annotated". Instead, you need to describe the shape, position of this area, and the physical properties that enable it to complete the task using descriptive language. Additionally, neither your thinking process nor output content shall include any coordinate values.

\section*{Thinking Steps}
1. What is the current task?  
   Analysis: The task requires identifying or locating which part of the object serves a specific purpose. This requires understanding the object’s function and combining it with image information for judgment.  

2. Where is the target object in the image, and what is its current posture (e.g., upright, lying flat, occluded) or state (e.g., microwave is closed, gripper is open)?  
   Analysis: Describe the position and state of the target object in the image to lay the foundation for subsequent functional reasoning.  

3. What are the common operable areas of this object?  
   Analysis: Based on common-sense understanding of the object, deduce which area is typically used when performing the specific task.  

4. If it is a grasping task, is the target within the grasping range or is there sufficient operating space?  
   Analysis: Determine whether the task is a grasping task or a functional area identification task. If there is a clear robotic arm or gripper in the image, it is a grasping task—evaluate whether the operable area is accessible and has sufficient space. This step does not apply if it is not a grasping task.  

5. Finally, summarize and identify which operable area can enable the entire task.  
   Analysis: Conclude and confirm that, based on the above analysis, the identified area is indeed the correct choice for completing the task.

\subsection*{Provided User Query:}

\subsection*{Example 1:}
\textbf{User Query:} Please indicate a possible affordance area of the baseball bat that can be used for holding  
\textbf{Output:}  
First, I need to determine the current task. The task is to identify the functional area on a baseball bat that is used for "holding." This requires understanding the bat's purpose as a tool and recognizing the specific part designed for gripping.  
Next, I need to analyze the object's location, posture, and current state within the image. In the picture, the baseball bat is on the left side, positioned vertically with no obvious obstructions.  
Then, I'll analyze the common operable areas of this object. Based on common knowledge, a baseball bat is used for hitting a ball. To swing effectively, a person typically holds the narrower, bottom part of the bat, which is the handle. This handle area is specifically designed for gripping.  
I must also determine if this is a grasping task and if the object is within grasping range or has sufficient operating space. Based on my analysis, this is not a grasping task but a "functional area identification" task. Therefore, grasping range and operating space are not applicable.  
Finally, I need to summarize and point out which operable area can accomplish the entire task. By holding the bat by its handle, a person can gain a stable grip and control it for the subsequent swinging motion. Therefore, the handle is the correct area on the baseball bat to complete this task.

\subsection*{Example 2:}
\textbf{User Query:} <Imagehere> Please indicate a possible affordance area of the bench that can be used for sitting  
\textbf{Output:}  
First, I need to determine the current task. The task is to identify the functional area on a bench that is used for "sitting." This requires understanding the function of a bench as a piece of furniture that provides a resting area for people or robots.  
Next, I need to analyze the object's location, posture, and current state within the image. There is a bench in the center of the frame. It's positioned horizontally and steadily, with no obvious obstructions. Its structure includes a seating surface and four support legs.  
Then, I will analyze the common operable areas of this object. The primary function of a bench is to provide a place to sit. Therefore, its horizontal seating surface is the specific operable area designed for this purpose. This area is typically flat and stable to support the weight of a person or object.  
I must also determine if this is a grasping task and if the object is within grasping range or has sufficient operating space. Based on my analysis, this is not a grasping task, but a "functional area identification" task. I need to understand and recognize the area's "affordance" for sitting. Therefore, evaluating grasping range or operating space is not necessary.  
Finally, I need to summarize and point out which operable area can accomplish the entire task. Based on the analysis above, the seating surface of the bench is the affordance area for sitting. Locating and indicating this area completes the task.

\end{tcolorbox}

\subsection{\textbf{Appendix B}  }
\begin{figure}[!t]
    \centering
    \includegraphics[width=0.98\linewidth]{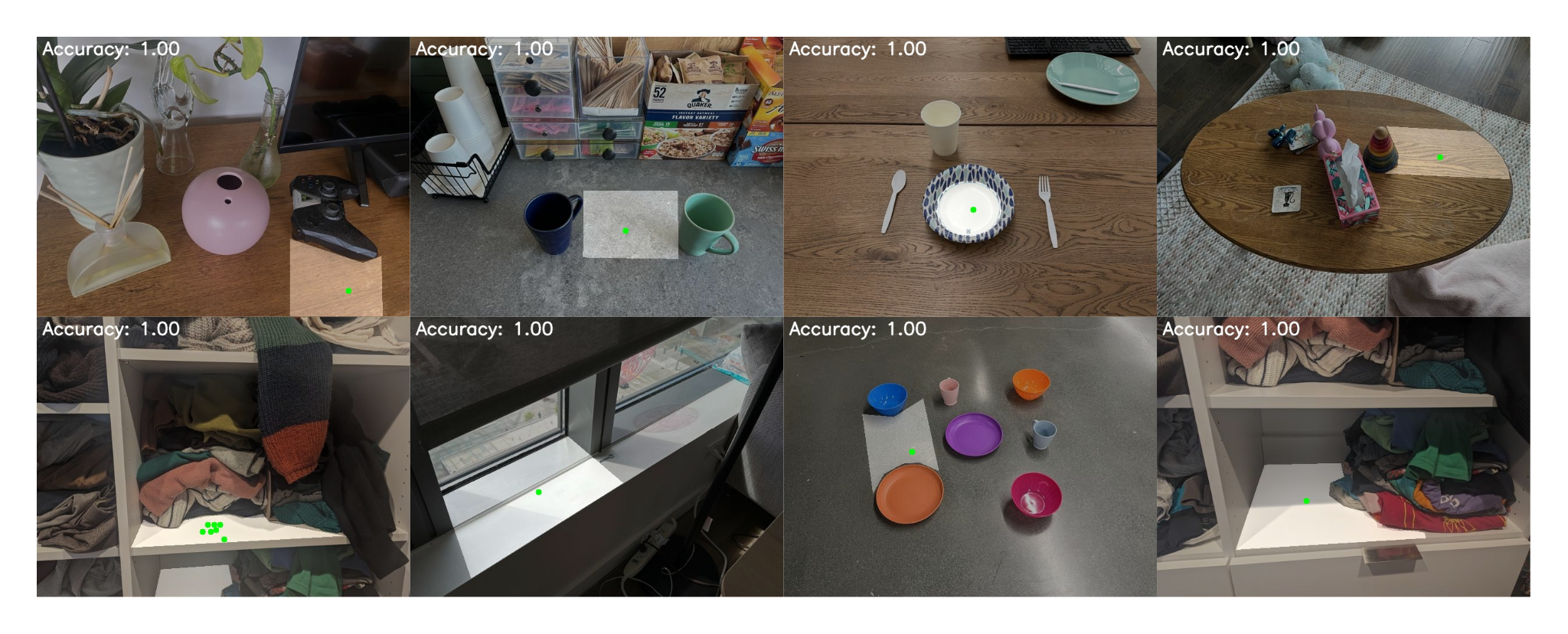}
    \caption{
        Partial results on the Where2Place-bench.
    }
    \label{fig:data}
\end{figure}

\begin{figure}[!t]
    \centering
    \includegraphics[width=0.98\linewidth]{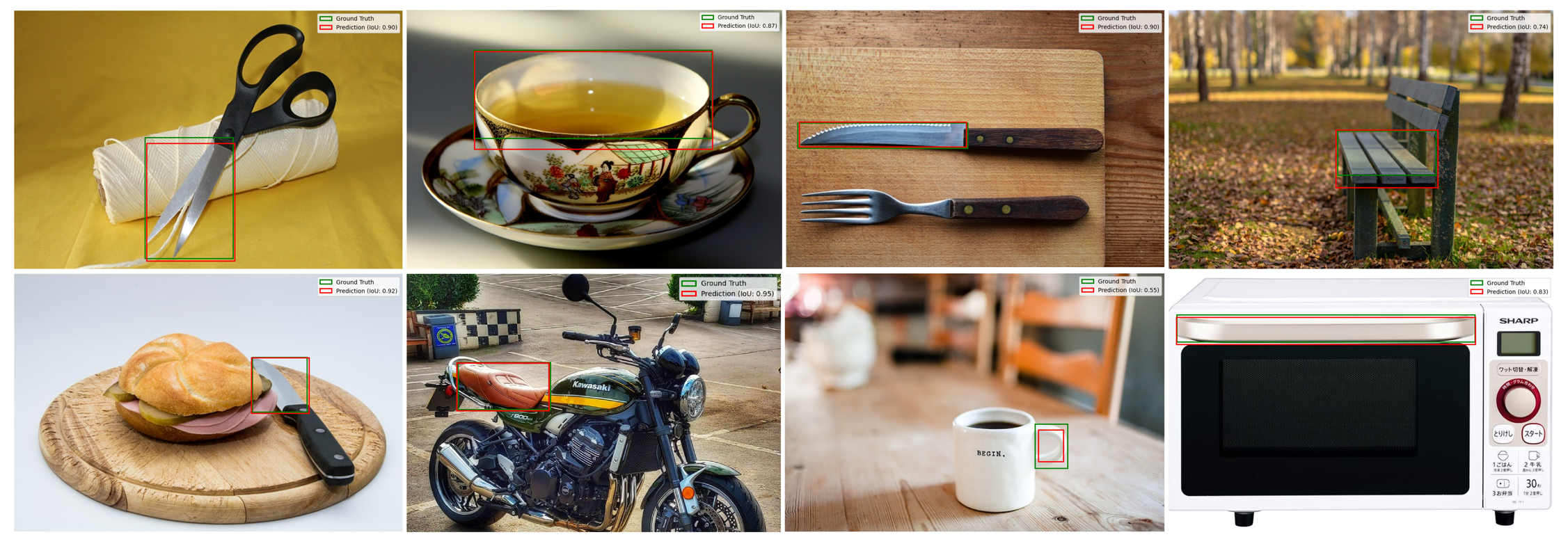}
    \caption{
        Partial results on the modified Sharerobot-bench.
    }
    \label{fig:data}
\end{figure}

\begin{figure}[!h]
    \centering
    \includegraphics[width=0.98\linewidth]{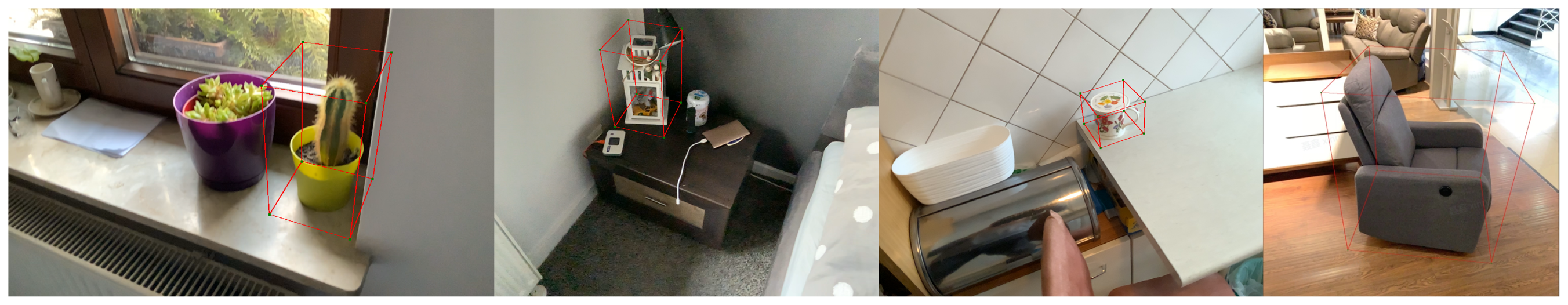}
    \caption{
        Partial results on the iFlyBot-3DGrounding-bench.
    }
    \label{fig:data}
\end{figure}

\begin{figure}[!h]
    \centering
    \includegraphics[width=0.98\linewidth]{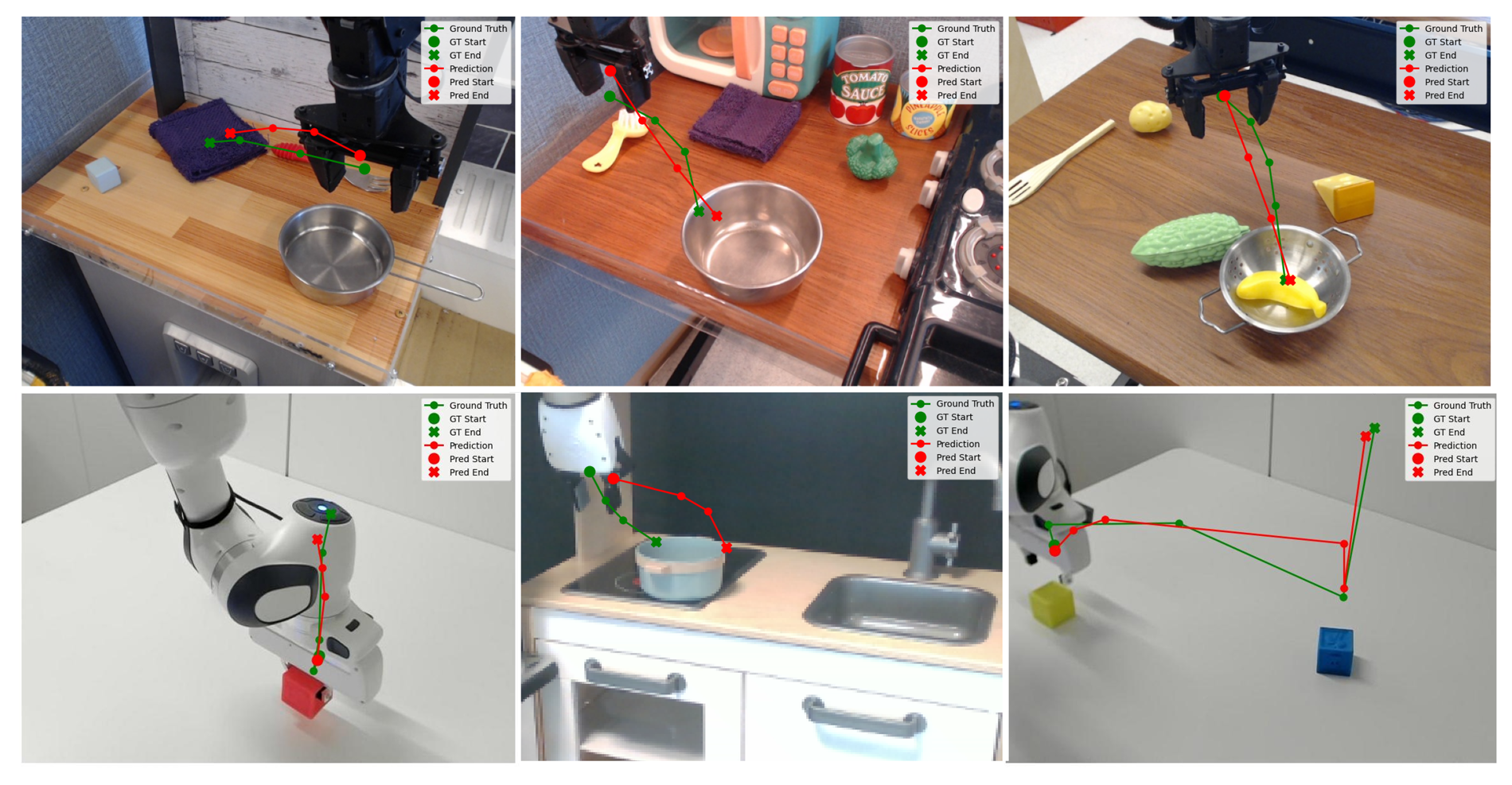}
    \caption{
        Partial results on the Sharerobot-bench.
    }
    \label{fig:data}
\end{figure}

\begin{figure}[!h]
    \centering
    \includegraphics[width=0.98\linewidth]{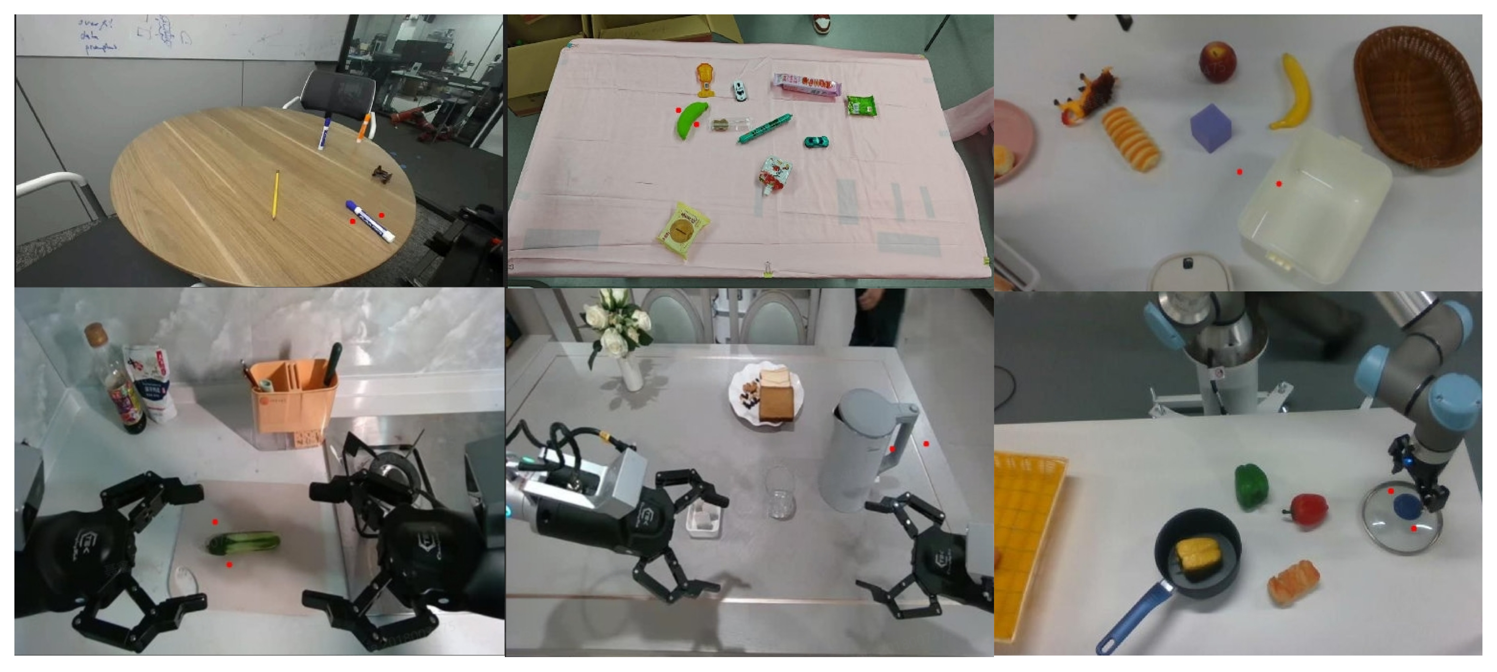}
    \caption{
        Partial results on the iFlyBot-GraspPose-bench.
    }
    \label{fig:data}
\end{figure}

\begin{figure}[!t]
    \centering
    \includegraphics[width=0.98\linewidth]{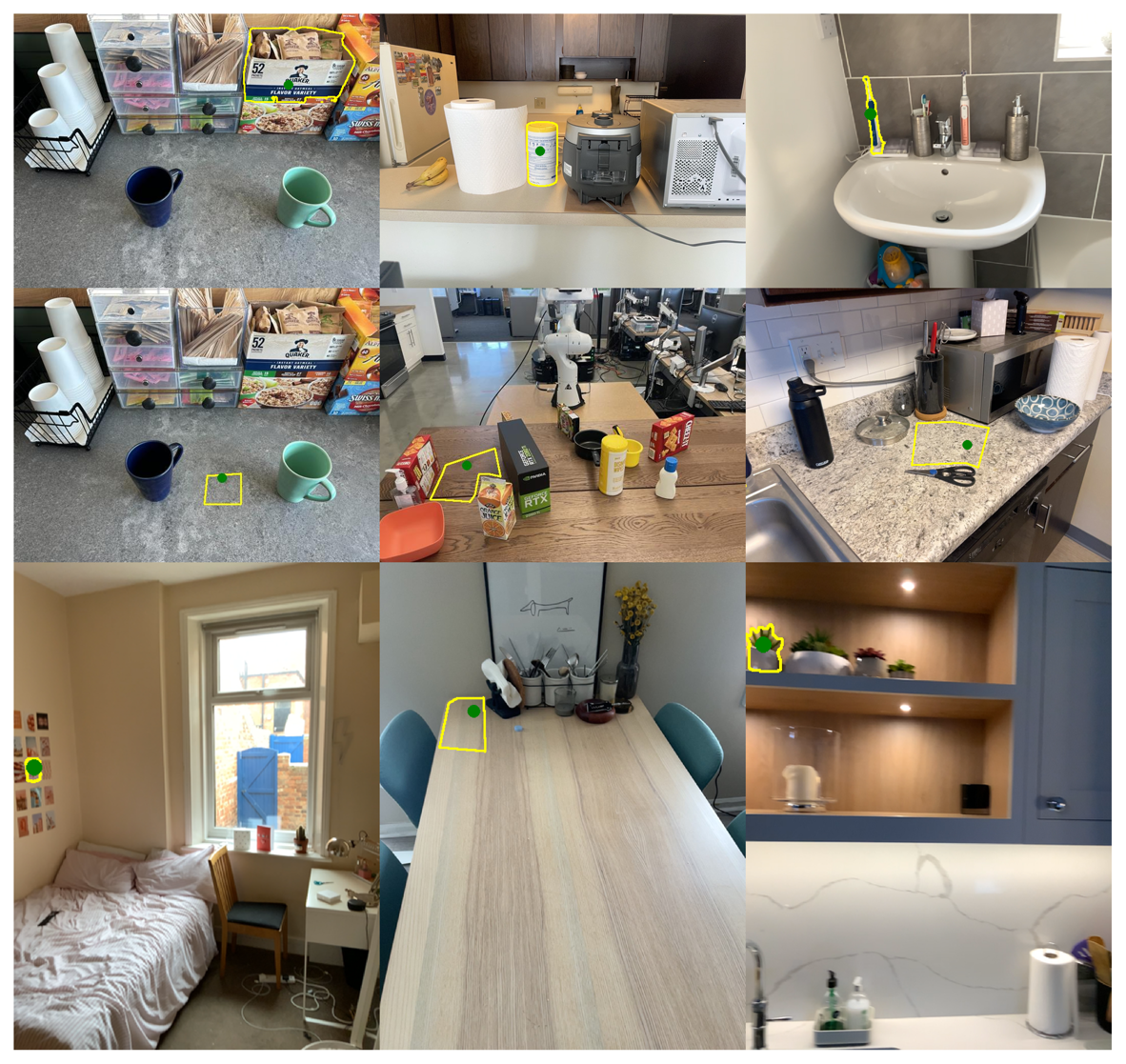}
    \caption{
        Partial results on the RefSpatial-bench.
    }
    \label{fig:data}
\end{figure}

\clearpage


\end{document}